\documentclass{article} 
\usepackage{iclr2020_conference,times}
\usepackage{tabularx}

\usepackage{amsmath,amsfonts,bm}

\def\eqref#1{equation~\ref{#1}}

\def\1{\bm{1}}

\DeclareMathAlphabet{\mathsfit}{\encodingdefault}{\sfdefault}{m}{sl}
\SetMathAlphabet{\mathsfit}{bold}{\encodingdefault}{\sfdefault}{bx}{n}

\def\sD{{\mathbb{D}}}

\def\sV{{\mathbb{V}}}

\usepackage{booktabs}
\usepackage{tipa}

\usepackage{hyperref}
\usepackage{url}

\title{{Combining Pretrained High-Resource Embeddings and Subword Representations for Low-Resource Languages}}

\iclrfinalcopy
\author{Machel Reid, Edison Marrese-Taylor \& Yutaka Matsuo\\
Graduate School of Engineering\\
The University of Tokyo\\
Tokyo, Japan \\
\texttt{machelreid2004@gmail.com, \{emarrese,matsuo\}@weblab.t.u-tokyo.ac.jp} \\
}

\begin{document}

\maketitle

\begin{abstract}
The contrast between the need for large amounts of data for current Natural Language Processing (NLP) techniques, and the lack thereof, is accentuated in the case of African languages, most of which are considered low-resource.
To help circumvent this issue, we explore techniques exploiting the qualities of morphologically rich languages (MRLs), while leveraging pretrained word vectors in well-resourced languages. In our exploration, we show that a meta-embedding approach combining both pretrained and morphologically-informed word embeddings performs best in the downstream task of Xhosa-English translation.

\end{abstract}
\section{Introduction \& Related Work}
Learning distributed representations of words \citep{mikolov2013distributed,pennington-etal-2014-glove} has found great success in many NLP tasks. 
However, the need for large amounts of data in low-resource African languages is a significant drawback of the current methods for learning word embeddings.
 \citet{shikali2019better} proposed learning Swahili word embeddings by taking morphemes into account, while \citet{grave2018learning} developed word embeddings for 157 languages, including multiple low-resource languages.

Learning of cross-lingual word embeddings \citep{DBLP:journals/corr/Ruder17} and projecting monolingual word embeddings to a single cross-lingual embedding space \citep{artetxe2018acl} have also been proposed to help learn embeddings for low-resource languages.  

In recent years, definitions and context words have been used to learn representations of nonce words (a word created for a single occasion to solve an immediate problem of communication), with \citet{doi:10.1111/cogs.12481} proposing learning nonce words by summing the representations of context words, obtaining representations highly correlated with human
judgements in terms of similarity. Additionally, \citet{TACL711} propose using word definitions to learn nonce word representations.

Our contributions are twofold: (1) Inspired by the work above, we explore techniques for combining pretrained high resource vectors and subword representations to produce better word embeddings for low-resource MRLs, and (2) we develop a new dataset, \textit{XhOxExamples}, made up of 5K Xhosa-English examples, that we collected from the \citet{oxford}.



\section{Approach}

Our approach assumes the existence of a vocabulary in our low-resource language $\sV = \{v_1,\dots,v_T\}$, the corresponding translations of the words in $\sV$ in a high-resource language, referred to as $\sD = \{d_1, \dots, d_T\}$, and a pretrained embedding matrix for the high resource language $E_{HR}$. $\sD$  
can either be comprised by individual words,
in the case the word in the low-resource language can be accurately mapped to a single word in the high-resource language (e.g. indoda $\rightarrow$ man)\footnote{All examples are from Xhosa to English}; 
or by a sequence of words in the case that the word in the low-resource language maps to more than one word in the high-resource language (e.g. bethuna $\rightarrow$ [listen, everyone]). Concretely, our objective is to use both $\sV$ and $\sD$ to produce vector representations for each word in $\sV$.

To leverage the high-resource language, we embed the atomic elements of $\sD$ in $E_{HR}$ and map the resulting vectors to the corresponding word in $\sV$. In the case of $d_i$ being a sequence, we take a similar approach to \citet{doi:10.1111/cogs.12481} and sum the normalized word vectors for each word in $d_i$ to produce a word representation for the word $v_i$. We refer to the resulting embedding matrix as $E_{V}$. Additionally, we pretrain another embedding matrix $E_{M}$ on a corpus in our low resource language using subword information to capture similarity correlated with the morphological nature of words \citep{DBLP:journals/corr/BojanowskiGJM16}.

We experiment with the following 4 methods to initialize the word embeddings for our downstream task\footnote{Note that the vocabulary for our downstream task might differ from $\sV$.}:
\begin{itemize}
    \item \textbf{\textit{Xh}Pre} - Initialization with $E_{V}$. Words not present in $E_V$ are initialized with $E_{M}$.
    \item \textbf{\textit{Xh}Sub} - Initialization with $E_{M}$ only.
    \item \textbf{VecMap} - We learn cross-lingual word embedding mappings by taking two sets of monolingual word embeddings, $E_{V}$ and $E_{M}$, and mapping them to a common space following \citet{artetxe2018acl}.
    \item \textbf{\textit{Xh}Meta} - We compute meta-embeddings for every word $w_i$ by taking the mean of $E_{V}(w_i)$ and $E_{M}(w_i)$, following \citet{coates-bollegala-2018-frustratingly}. Words not present in $E_V$ are associated with an UNK token and its corresponding vector.
\end{itemize}

\begin{table}
\begin{center}
\begin{tabular}{l|ll|llr} 
\toprule
& \multicolumn{2}{c|}{Bible}
& \multicolumn{2}{c}{XhOxExamples}\\
Aspect & Xhosa  & English &  Xhosa & English\\
\midrule
Sentence Length (Mean$\pm$Std)  & 18.44$\pm$7.74 & 29.43$\pm$12.61   &
5.16$\pm$1.89 &
8.42$\pm$3.04 \\
Total \# of Tokens & 573571 & 915531 & 25978 & 42400\\
\# of Sentences & \multicolumn{2}{c|}{31102}
& \multicolumn{2}{c}{5033}\\
Train/Dev/Test Ratio & \multicolumn{2}{c|}{70/20/10}
& \multicolumn{2}{c}{70/20/10}\\
\bottomrule
\end{tabular}
\end{center}
\caption{\label{tab:stats}Statistics regarding the Bible corpus and \textit{XhOxExamples} }
\end{table}
\begin{table}
\begin{center}
\begin{tabular}{l|ll|lr} 
\toprule
& \multicolumn{2}{c|}{Bible}
& \multicolumn{2}{c}{XhOxExamples}\\
Model & BLEU  & BEER & BLEU-4 & BEER\\
\midrule
Random Initializaton  &  21.79    & 21.84 & 16.08 & 25.30 \\
VecMap          &    22.46     &  22.03 & 16.38 & 25.42 \\
\textit{Xh}Sub &  24.65 & 22.79 & 17.37 & 26.04 \\
\textit{Xh}Pre       & 27.67     & 22.40 & 17.06 & 25.70\\
\midrule
\textit{Xh}Meta        & \textbf{29.09}  & \textbf{23.33}  & \textbf{17.77} & \textbf{26.44}   \\
\bottomrule
\end{tabular}
\end{center}
\caption{\label{tab:table}Results on the test set of the Xhosa-English Bible Corpus and \textit{XhOxExamples}}
\end{table}

\section{Experiments \& Results}

To test the performance of the 4 different embedding configurations, we both train and evaluate using machine translation (MT) as our downstream task on the English and Xhosa versions of the parallel Bible corpus \citep{christodouloupoulos2015massively} and make use of two MT evaluation metrics, BEER \citep{stanojevic-simaan-2014-beer} and sentence-level BLEU \citep{papineni2002bleu}. We also fine tune and evaluate the above pretrained models on the \textit{XhOxExamples} dataset to evaluate performance after fine-tuning as shown on the right side of Table \ref{tab:table}. Statistics for both datasets can be seen in Table \ref{tab:stats}.

All models are trained using OpenNMT \citep{klein-etal-2017-opennmt}, with a 2-layer 128-dimensional BiGRU for encoding, and 2-layer 128-dimensional GRU \citep{Cho_2014} using a beam search algorithm with a beam size of 5 for decoding. Embedding matrix $E_V$ is initialized with the 840B-token version\footnote{\url{https://nlp.stanford.edu/projects/glove/}} of GloVe \citep{pennington-etal-2014-glove}, while $E_M$ is initialized with a \texttt{FASTTEXT} model \citep{DBLP:journals/corr/BojanowskiGJM16}, trained on a combination of \textit{XhOxExamples} and the Xhosa Bible Corpus.

As is evident from the results shown in Table \ref{tab:table}, \textit{Xh}Meta outperforms all other models in both benchmarks. We believe this is a consequence of the simple approach of meta-embedding, resulting in vectors being both similar to both the subword representations captured by $E_M$ and the pretrained vectors, capturing a sense of the ``meaning" of the word, demonstrating the necessity of both aspects in this context.
\section{Conclusion}
In this paper, we explore different ways of combining pretrained high-resource word vectors and subword representations to produce more meaningful word embeddings for low-resource MRLs. We show that both types of representations are important in the context of the low-resourced MRL, Xhosa, and hope that this research is able to assist others when doing NLP for African languages.
\\~\\

\large{\textbf{Acknowledgements}}
\normalsize

We are grateful to Jorge Balazs for his fruitful
corrections and comments. We also gratefully acknowledge the support of NVIDIA Corporation with the donation of the Titan Xp GPU used for this research.
\bibliography{iclr2020_conference}
\bibliographystyle{iclr2020_conference}

\end{document}